\documentclass[11pt]{article}

\usepackage{acl}
\usepackage{times}
\usepackage{latexsym}
\usepackage[T1]{fontenc}
\usepackage[utf8]{inputenc}
\usepackage{microtype}

\usepackage{graphicx}
\usepackage{booktabs}
\usepackage{amsmath,amssymb}
\usepackage{xcolor}
\usepackage{multirow}
\usepackage{url}
\usepackage{pgfplots}
\pgfplotsset{compat=1.18}
\usepackage{subcaption}
\usepackage{enumitem}
\usetikzlibrary{shapes.geometric, arrows.meta, positioning, backgrounds, fit}

\definecolor{accent1}{HTML}{2B6CB0}

\tikzset{
  fbox/.style={
    rectangle, rounded corners=1.5pt,
    draw=black!65, fill=accent1!18,
    minimum height=0.5cm, inner sep=3pt, outer sep=0pt,
    align=center, font=\rmfamily\scriptsize,
  },
  farr/.style={
    -{Stealth[length=2.5pt]}, semithick, color=black!75,
  },
  flbl/.style={
    font=\rmfamily\tiny, color=black!90, align=center,
  },
  fdiamond/.style={
    diamond, aspect=2, draw=black!65, fill=accent1!18,
    inner sep=1.5pt, font=\rmfamily\scriptsize, align=center,
  },
  fregion/.style={
    draw=black!45, dashed, rounded corners=3pt,
    fill=accent1!8, inner sep=6pt,
  },
  fbox-blue/.style={fbox, fill=blue!20, draw=blue!55},
  fbox-wide/.style={fbox, minimum width=2.2cm},
  farr-skip/.style={farr, dashed},
}

\usepackage{pifont}
\newcommand{\cmark}{\ding{51}}
\newcommand{\xmark}{\ding{55}}

\definecolor{trackquran}{RGB}{0,114,178}
\definecolor{trackhadith}{RGB}{213,94,0}
\definecolor{trackfiqh}{RGB}{0,158,115}

\definecolor{madhabHanafi}{RGB}{31,119,180}
\definecolor{madhabMaliki}{RGB}{255,127,14}
\definecolor{madhabShafii}{RGB}{44,160,44}
\definecolor{madhabHanbali}{RGB}{214,39,40}

\definecolor{tierBaseline}{RGB}{158,158,158}
\definecolor{tierArabic}{RGB}{255,193,7}
\definecolor{tierMid}{RGB}{100,181,246}
\definecolor{tierFrontier}{RGB}{129,199,132}

\title{IslamicMMLU:\\A Benchmark for Evaluating LLMs on Islamic Knowledge}

\author{
  \textbf{Ali Abdelaal, Mohammed Nader Al Haffar,} \\
  \textbf{Mahmoud Fawzi, Walid Magdy} \\
  The University of Edinburgh \\
  \texttt{\{A.Abdelaal, M.N.Al-haffar, M.F.G.Ibrahim\}@sms.ed.ac.uk} \\
  \texttt{wmagdy@inf.ed.ac.uk}
}

\begin{document}

\maketitle
% \vspace*{-5em}

\begin{abstract}
Large language models are increasingly consulted for Islamic knowledge, yet no comprehensive benchmark evaluates their performance across core Islamic disciplines.
We introduce IslamicMMLU, a benchmark of 10,013 multiple-choice questions spanning three tracks: Quran (2,013 questions), Hadith (4,000 questions), and Fiqh (jurisprudence, 4,000 questions). Each track is formed of multiple types of questions to examine LLMs capabilities handling different aspects of Islamic knowledge. The benchmark is used to create the IslamicMMLU public leaderboard for evaluating LLMs, and we initially evaluate 26 LLMs, where their averaged accuracy across the three tracks varied between 39.8\% to 93.8\% (by Gemini 3 Flash). The Quran track shows the widest span (99.3\% to 32.4\%), while the Fiqh track includes a novel madhab (Islamic school of jurisprudence) bias detection task revealing variable school-of-thought preferences across models. Arabic-specific models show mixed results, but they all underperform compared to frontier models. The evaluation code and leaderboard are made publicly available.
\end{abstract}

\section{Introduction}
\label{sec:introduction}

Large language models (LLMs) are increasingly consulted for information about Islamic topics by millions of users worldwide \cite{mubarak2025islamiceval}. Islamic knowledge spans multiple interconnected disciplines (e.g., Quranic studies, Hadith sciences, and Islamic jurisprudence), each requiring distinct scholarly expertise and evaluation methodologies. Yet no comprehensive benchmark evaluates LLM performance across these core domains. This gap can lead to religious misinformation affecting daily practice for Muslims who may lack easy access to qualified scholars~\citep{naous2024beer,fawzi2026fabricatingholiness}.

Several recent efforts address aspects of Islamic NLP (e.g., hallucination detection, cultural competence, and jurisprudential QA), but none provides comprehensive cross-discipline evaluation. The MMLU paradigm~\citep{hendrycks2021mmlu} has been extended to legal~\citep{hijazi2024arablegaleval}, cultural~\citep{huang2024acegpt}, and general Arabic~\citep{koto2024arabicmmlu} domains, but to our knowledge no systematic Islamic knowledge evaluation benchmark exists.

Islamic knowledge evaluation faces unique challenges across all three disciplines. Quranic evaluation requires precise textual recall across 6,236 verses (\textit{Ayahs}). Hadith evaluation demands source authentication across authentic canonical collections, textual completion, and reliability grading of authenticity. Fiqh evaluation confronts the inherent pluralism of Islamic jurisprudence, where the same act may receive different rulings across multiple valid schools of thought (\textit{madhahib}).``Correct/Incorrect'' evaluation is fundamentally insufficient for fiqh, as a model penalised for answering according to one school when the benchmark expected another is not exhibiting error but reflecting the genuine plurality of the tradition.

We introduce IslamicMMLU with the following contributions:

\begin{enumerate}[noitemsep,topsep=0pt]
    \item \textbf{IslamicMMLU benchmark:} 10,013 multiple-choice questions spanning three tracks with 12 task types that evaluate different facets of Islamic knowledge.

    \item \textbf{Fiqh track with madhab bias detection:} A novel evaluation task measuring implicit school-of-thought bias in LLMs.

    \item \textbf{Evaluation of 26 LLMs:} Comprehensive evaluation across all three tracks with bootstrap confidence intervals and statistical significance testing.

    \item \textbf{IslamicMMLU leaderboard\footnote{\url{https://huggingface.co/spaces/islamicmmlu/leaderboard}}:} Evaluation code and results released publicly via HuggingFace, with a live leaderboard supporting community evaluation of new models.
\end{enumerate}

\section{Background and Related Work}
\label{sec:related}

\subsection{Islamic Knowledge Domains}

Islam, the world's second-largest religion, centres on three interconnected scholarly traditions that underpin religious practice and daily life.
The \textbf{Quran} is the central scripture of Islam, comprising 114 chapters (\textit{surahs}) and 6,236 verses (\textit{ayahs}). Muslims believe it to be the verbatim word of God as revealed to Muhammad, the prophet of Islam. The Quran is recited in the five daily prayers that structure Muslim worship, and is among the most widely memorised religious texts in the world \cite{abokhodair2020holytweets}.

The \textbf{Hadith} literature records the sayings, actions, and tacit approvals of the Prophet Muhammad~\citep{fawzi2025prophetsaidso}. It consists of two parts: 1) \textit{isnad} which is the chain of narrators who narrated the quote of the prophet till the time it has been written, and 2) \textit{matn}, which is the main content of the quote itself. Islam has two main branches. The Sunni branch which is followed by the majority of Muslims and the Shia branch.
Six canonical collections of Hadith are widely accepted by Sunni Muslims as the most reliable source of Hadith,  which are Bukhari, Muslim, Abu Dawud, Tirmidhi, Nasa'i, and Ibn Majah. %
Hadith sciences classify narrations by authenticity according to how evident it has been said by the prophet. There are four main authenticity levels: \textit{sahih} (authentic), \textit{hasan} (good), \textit{da`if} (weak), and \textit{mawdu`} (fabricated). Shia Muslims use other books.

Both Quran and Hadith are the main sources of Islamic jurisprudence (in Arabic \textbf{Fiqh}).   There are four major Sunni schools (\textit{madhab}), namely Hanafi, Maliki, Shafi'i, and Hanbali, developed over centuries and are recognised as equally valid. They often agree but may reach different rulings on the same question, creating a legitimate pluralism that standard single-answer evaluation paradigms cannot capture. The Shia disagreement about Hadith may lead to to some variation in Fiqh. Nevertheless, it still agrees with Sunni in many rulings regarding life and worship.

\subsection{MMLU and LLM Evaluation}

The MMLU benchmark~\citep{hendrycks2021mmlu} established MCQ-based knowledge evaluation as a standard paradigm, with 15,908 questions across 57 subjects. MMLU-Pro~\citep{wang2024mmlupro} extended this to harder questions with ten options. Domain-specific adaptations have emerged for law, medicine, and cultural knowledge, establishing MCQ evaluation as the reproducible, scalable framework for cross-model comparison. HELM~\citep{liang2023helm} provides a holistic evaluation framework spanning multiple scenarios and metrics.

\subsection{Arabic NLP Benchmarks}

ArabicMMLU~\citep{koto2024arabicmmlu} is the most direct comparator: 14,575 native Arabic questions across 40 subjects sourced from school exams in eight Arab countries, with quality validation on 100 samples achieving 96\% accuracy. It includes Islamic studies as a single category with basic questions without the domain depth needed to examine knowledge in different nuances of Islam. ACVA, introduced with AceGPT~\citep{huang2024acegpt}, offers 8,000+ true/false questions on Arabic cultural values but explicitly excludes religious jurisprudence. AlGhafa~\citep{almazrouei2023alghafa} provides a general Arabic evaluation benchmark but does not cover Islamic knowledge domains in depth. ILMAAM~\citep{nacar2025ilmaam} provides a culturally aligned Arabic MMLU variant. ArabLegalEval~\citep{hijazi2024arablegaleval} evaluates LLMs on Saudi secular law with 10,000+ MCQs for domain-specific evaluation. A recent survey catalogues over 40 Arabic benchmarks and identifies gaps in religious domain coverage~\citep{alzubaidi2025arabicsurvey}.

\subsection{Cultural Bias in LLMs}

\citet{naous2024beer} demonstrate that LLMs default to Western cultural norms even when prompted in Arabic. \citet{plaza2024divine} find differential treatment across faiths, with Islam facing elevated stereotyping and disproportionate refusal rates. \citet{abid2021} provide foundational evidence of persistent anti-Muslim bias in GPT-3. DLAMA~\citep{keleg2023dlama} reveals Western bias in multilingual knowledge probing.

\begin{table}[t]
    \centering
    \footnotesize
    \setlength{\tabcolsep}{1pt}
    \begin{tabular}{lccccccc}
        \toprule
        \textbf{Benchmark} & \textbf{Lang} & \textbf{\#Q} & \textbf{Tracks} & \textbf{Bias} & \textbf{Pub.} & \textbf{Val.} \\
        \midrule
        ArabicMMLU       & ar   & 14.6k & 1\textsuperscript{\dag} & \xmark & \cmark & 2 ann. \\
        ACVA             & ar   & 8k+   & 1   & \xmark & \cmark & --- \\
        IslamicEval '25  & ar   & ---   & 2   & \xmark & \cmark & task  \\
        PalmX            & ar   & ---   & 1   & \xmark & \cmark & --- \\
        FiqhQA           & ar   & 960   & 1   & \xmark & \cmark & synth \\
        Isl.LegalBench   & ar   & 718   & 1   & \xmark & \cmark & expert \\
        AlGhafa          & ar   & 10k+  & 0   & \xmark & \cmark & --- \\
        \midrule
        \textbf{IslamicMMLU} & \textbf{ar} & \textbf{10k} & \textbf{3} & \cmark & API & \textbf{expert} \\
        \bottomrule
    \end{tabular}
    \caption{Comparison with related benchmarks. \#Q = question count; Tracks = Islamic discipline count (\textsuperscript{\dag}ArabicMMLU has Islamic Studies as one of 40 subjects); Bias = madhab bias detection; Pub. = public access (\cmark\ = open, API = leaderboard-only); Val. = validation method.}
    \label{tab:comparison}
\end{table}

\subsection{Islamic and Religious NLP}

IslamicEval 2025~\citep{mubarak2025islamiceval} is the first shared task targeting LLM hallucinations in Islamic content, attracting 13 teams. The shared task offers solution to LLMs hallucination with Quran and Hadith. Their released dataset highlights the presence of hallucination in Islamic content by different LLMs, which further motivates the need for a benchmark that measures LLMs knowledge of Islamic content.

PalmX~\citep{alwajih2025palmx} addresses cultural competence including an Islamic culture subtask. QIAS~\citep{bouchekif2025qias} covers Islamic inheritance reasoning. FiqhQA~\citep{fiqhqa2025} introduces 960 question-answer pairs categorised by school but relies on synthetically generated answers from a single LLM, raising concerns about circular evaluation. IslamicLegalBench~\citep{elmahjub2026islamiclegalbench} provides 718 manually curated instances across seven schools of jurisprudence, measuring legal reasoning depth. QuranQA~\citep{malhas2022qrcd} establishes reading comprehension benchmarks for Quranic text. IslamTrust~\citep{lahmar2025islamtrust} evaluates alignment with consensus Islamic ethics, achieving only 66.5\% alignment with GPT-4.

Despite these initiatives, a significant gap remains for a comprehensive benchmark and leaderboard that systematically evaluates LLM performance across diverse Islamic domains and measures potential bias toward various schools of Fiqh. IslamicMMLU is designed to address this gap (see Table \ref{tab:comparison}).

\section{Data Preparation Methodology}
\label{sec:methodology}

IslamicMMLU follows the MMLU paradigm: 4-option MCQ, standardised Arabic prompts, and zero-shot evaluation. All questions are generated from native Arabic content sourced from authoritative Islamic texts (not translated from English), programmatic question generation with quality verification, and domain-expert review when needed. IslamicMMLU consists of three main tracks, each differ in source material, question type design, and domain-specific challenges. Table~\ref{tab:benchmark-overview} summarises the three tracks.

\begin{table}[t]
    \centering
    \footnotesize
    \setlength{\tabcolsep}{4pt}
    \begin{tabular}{lccc}
        \toprule
        & \textbf{Quran} & \textbf{Hadith} & \textbf{Fiqh} \\
        \midrule
        Questions     & 2,013  & 4,000  & 4,000  \\
        Task types    & 3      & 4      & 5      \\
        Sources       & 114 surahs & 6 collections & 1 encyclopedia \\
        Bias analysis & ---    & ---    & Madhab \\
        \bottomrule
    \end{tabular}
    \caption{IslamicMMLU benchmark overview. 10,013 questions from 3 tracks across 12 task types.}
    \label{tab:benchmark-overview}
\end{table}

\subsection{Quran Track}
\label{subsec:quran-track}

The Quran track comprises 2,013 four-way multiple-choice questions covering all 114 surahs ($S$). Sourced from standard Arabic Quranic text, the dataset underwent normalization (e.g., removal of diacritics and kashida; unification of Alef and Ya forms) following \cite{ArabicIR}. To ensure validity, all questions pass a programmatic uniqueness constraint to prevent duplicate candidates or distractors matching the ground truth.

\textit{Verse (Ayah) Count} (114 questions, 6\%): This task evaluates factual recall of surah lengths. Distractors are selected via a structured strategy: (1) ground truth; (2) subsequent $S$ count; (3) preceding $S$ count; and (4) a random $S$ count. If preceding or subsequent counts match the target, we apply an incremental offset ($\pm 2$) to maintain four unique candidates.

\textit{Surah Identification} (833 questions, 41\%): Models must identify the source $S$ of a given verse. To prioritize contextual reasoning over keyword matching, we only include verses with a Jaccard similarity index $J > 0.4$ relative to at least one verse in a different $S$. We employ hard-negative mining to select: (1) the ground truth $S$; (2) the nearest-neighbor $S$ containing the most lexically similar verse; and (3-4) two randomly sampled surahs.

\textit{Verse Retrieval} (1,066 questions, 53\%): This task requires mapping a verse index $V$ to its textual content within Surah $S$. To increase linguistic complexity, $V$ is stochastically represented as either a numeric digit or its Arabic textual equivalent. We filter for verses with $J > 0.4$ across the corpus and construct adversarial choices: (1) the ground truth (max 10 words); (2) the most lexically similar verse in the corpus; (3) a positional distractor at an offset of $\pm 2$ from $V$ within $S$; and (4) a "null response" indicating "No verse at this position." The latter acts as the correct answer for \emph{trap} questions where $V$ exceeds the total verse count of $S$, specifically testing for model hallucination and boundary awareness.

\subsection{Hadith Track}
\label{subsec:hadith-track}

The Hadith track comprises 4,000 questions spanning the six canonical Sunni collections (\textit{Kutub al-Sittah})
The dataset employs a deterministic generation pipeline with fixed random seeds for full reproducibility.

\paragraph{Preprocessing.} Raw hadith texts undergo the same standard Arabic text normalisation applied to Quran text;
Critically, the \textit{isnad} (chain of narrators) is trimmed from each hadith before question generation to prevent models from trivially identifying the source collection from narrator names rather than demonstrating genuine content knowledge, forcing models to identify collections from the \textit{matn}, the main content alone. Cross-collection filtering removes near-identical narrations appearing in multiple collections with different attributions. %

\paragraph{Question Types.} The track tests four facets of Hadith knowledge:

\textit{Source Identification} (1,000 questions): Given a hadith text with the isnad removed, identify which canonical collection it belongs to. This is applied to Hadith that appeared in only one of the six books. Distractors are other collections.

\textit{Cloze Completion} (1,000 questions): Fill in a missing keyword from a hadith text. Target words are selected via IDF weighting to ensure the blank tests substantive vocabulary rather than common particles. Distractors are synonyms to the hidden word generated by an LLM (GPT4o).

\textit{Chapter Classification} (1,000 questions): Given a hadith from a given collection, identify which chapter (topic) within the collection it belongs to (e.g., Prayer, Fasting). This is the most challenging type, as it requires mapping narrative content to taxonomic labels rather than keyword matching.

\textit{Authenticity Grading} (${\sim}$1,000 questions): Assess the reliability grade of a hadith \{authentic, good, weak, fabricated\}. This carries the highest practical stakes, as incorrect grading can affect which narrations are used for deriving Islamic rulings. Items where narrations appear near-identically across collections with different grades are filtered to reduce ambiguity. We split this set of questions between the four levels of authenticity, where fabricated ones were taken from a set of Arabic poems as distractors.

\subsection{Fiqh Track}
\label{subsec:fiqh-track}

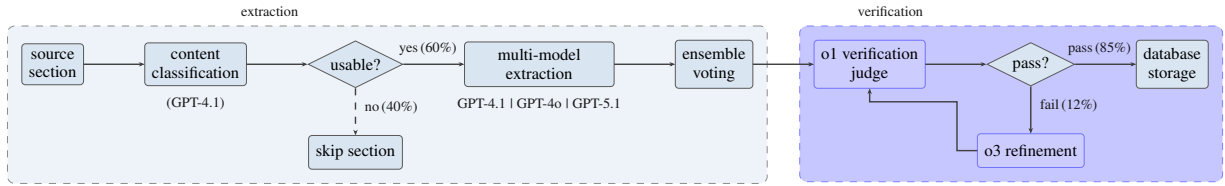
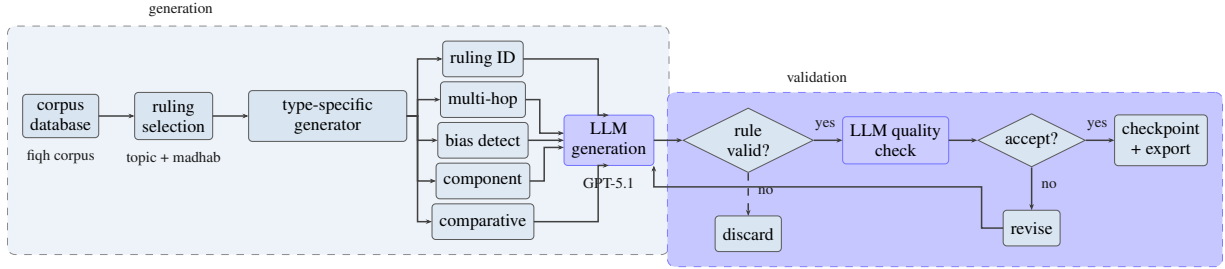
\begin{figure*}[!ht]
\centering
\vspace{-3.5em}

\begin{subfigure}[t]{\textwidth}
    \centering
    \resizebox{\textwidth}{!}{%
        \begin{tikzpicture}[node distance=0.5cm and 0.9cm]

            \node[fbox] (source) {source\\section};
            \node[fbox, right=of source] (classify) {content\\classification};
            \node[flbl, below=1pt of classify] {(GPT-4.1)};
            \node[fdiamond, right=of classify] (check1) {usable?};
            \node[fbox-wide, right=of check1] (extract) {multi-model\\extraction};
            \node[flbl, below=1pt of extract] {GPT-4.1\;\textbar\;GPT-4o\;\textbar\;GPT-5.1};
            \node[fbox, right=of extract] (voting) {ensemble\\voting};

            \node[fbox-blue, right=of voting] (verify) {o1 verification\\judge};
            \node[fdiamond, right=of verify] (check2) {pass?};
            \node[fbox, right=of check2] (store) {database\\storage};

            \node[fbox, below=0.7cm of check1] (skip) {skip section};

            \node[fbox-blue, below=0.7cm of check2] (refine) {o3 refinement};

            \draw[farr] (source) -- (classify);
            \draw[farr] (classify) -- (check1);
            \draw[farr] (check1) -- node[flbl, above, pos=0.4] {yes\,(60\%)} (extract);
            \draw[farr-skip] (check1) -- node[flbl, right, pos=0.4] {no\,(40\%)} (skip);
            \draw[farr] (extract) -- (voting);
            \draw[farr] (voting) -- (verify);
            \draw[farr] (verify) -- (check2);
            \draw[farr] (check2) -- node[flbl, above, pos=0.4] {pass\,(85\%)} (store);
            \draw[farr] (check2) -- node[flbl, right, pos=0.4] {fail\,(12\%)} (refine);
            \draw[farr] (refine.west) -- ++(-0.3,0) |- ([yshift=-0.2cm]verify.south) -- (verify.south);

            \begin{pgfonlayer}{background}
                \node[fregion, fit=(source)(classify)(check1)(extract)(voting)(skip),
                    label={[flbl]above left:extraction}] {};
                \node[fregion, fit=(verify)(check2)(refine)(store),
                    label={[flbl]above left:verification},
                    fill=blue!22, draw=blue!50] {};
            \end{pgfonlayer}

        \end{tikzpicture}%
    }%
    \caption{Multi-agent extraction pipeline. Three models extract in parallel with ensemble voting, followed by o1 verification and o3 refinement for failed extractions.}
    \label{fig:app-extraction}
\end{subfigure}

\vspace{12pt}

\begin{subfigure}[t]{\textwidth}
    \centering
    \resizebox{\textwidth}{!}{%
        \begin{tikzpicture}[node distance=0.4cm and 0.5cm]

            \node[fbox] (db) {corpus\\database};
            \node[flbl, below=1pt of db] {fiqh corpus};

            \node[fbox, right=of db] (select) {ruling\\selection};
            \node[flbl, below=1pt of select] {topic + madhab};

            \node[fbox-wide, right=of select] (typegen) {type-specific\\generator};

            \node[fbox, fill=accent1!15, right=0.5cm of typegen, minimum width=1cm, yshift=0.8cm]
            (t1) {ruling ID};
            \node[fbox, fill=accent1!15, below=2pt of t1, minimum width=1cm]
            (t2) {multi-hop};
            \node[fbox, fill=accent1!15, below=2pt of t2, minimum width=1cm]
            (t3) {bias detect};
            \node[fbox, fill=accent1!15, below=2pt of t3, minimum width=1cm]
            (t4) {component};
            \node[fbox, fill=accent1!15, below=2pt of t4, minimum width=1cm]
            (t5) {comparative};

            \node[fbox-blue, right=0.5cm of t3] (llmgen) {LLM\\generation};
            \node[flbl, below=1pt of llmgen] {GPT-5.1};

            \node[fdiamond, right=0.4cm of llmgen] (rulecheck) {rule\\valid?};
            \node[fbox-blue, right=0.4cm of rulecheck] (llmcheck) {LLM quality\\check};
            \node[fdiamond, right=0.4cm of llmcheck] (accept) {accept?};
            \node[fbox, right=0.4cm of accept] (export) {checkpoint\\+ export};

            \node[fbox, below=0.6cm of rulecheck] (discard) {discard};

            \node[fbox, below=0.6cm of accept] (revise) {revise};

            \draw[farr] (db) -- (select);
            \draw[farr] (select) -- (typegen);

            \draw[farr] (typegen.east) -- ++(0.15,0) |- (t1.west);
            \draw[farr] (typegen.east) -- ++(0.15,0) |- (t2.west);
            \draw[farr] (typegen.east) -- ++(0.15,0) |- (t3.west);
            \draw[farr] (typegen.east) -- ++(0.15,0) |- (t4.west);
            \draw[farr] (typegen.east) -- ++(0.15,0) |- (t5.west);

            \draw[farr] (t1.east) -| ([xshift=-0.15cm]llmgen.north) -- (llmgen.north);
            \draw[farr] (t2.east) -- ++(0.2,0) |- ([yshift=0.1cm]llmgen.west);
            \draw[farr] (t3.east) -- (llmgen.west);
            \draw[farr] (t4.east) -- ++(0.2,0) |- ([yshift=-0.1cm]llmgen.west);
            \draw[farr] (t5.east) -| ([xshift=-0.15cm]llmgen.south) -- (llmgen.south);

            \draw[farr] (llmgen) -- (rulecheck);
            \draw[farr] (rulecheck) -- node[flbl, above, pos=0.4] {yes} (llmcheck);
            \draw[farr-skip] (rulecheck) -- node[flbl, right, pos=0.4] {no} (discard);
            \draw[farr] (llmcheck) -- (accept);
            \draw[farr] (accept) -- node[flbl, above, pos=0.4] {yes} (export);
            \draw[farr] (accept) -- node[flbl, right, pos=0.4] {no} (revise);

            \draw[farr] (revise.west) -- ++(-0.3,0) |- ([yshift=-0.30cm]llmgen.south east) -- ([yshift=0cm]llmgen.south east);

            \begin{pgfonlayer}{background}
                \node[fregion,
                    fit=(db)(select)(typegen)(t1)(t5)(llmgen),
                    label={[flbl]above left:generation}] {};
                \node[fregion,
                    fit=(rulecheck)(llmcheck)(accept)(discard)(revise)(export),
                    label={[flbl]above left:validation},
                    fill=blue!22, draw=blue!50] {};
            \end{pgfonlayer}

        \end{tikzpicture}%
    }%
    \caption{Question generation pipeline. Rulings are routed to type-specific generators, producing candidates via GPT-5.1. Each question undergoes rule-based validation followed by LLM quality check before inclusion.}
    \label{fig:app-qgen}
\end{subfigure}

\caption{Fiqh track pipeline architectures. (a)~Extraction of structured rulings from al-Jaziri's source text. (b)~Generation and validation of benchmark questions from the structured corpus.}
\label{fig:app-pipelines}
\vspace{-1.5em}

\end{figure*}

The Fiqh track utilised the encyclopedia book "Jurisprudence According to the Four Schools" by `Abd al-Rahman al-Jaziri, commissioned by Al-Azhar University. The complete text spans 2,221 pages across 1,043 sections covering eight fiqh categories. We extract a structured corpus of 2,163 rulings across 797 topics using a multi-agent pipeline with adversarial verification  (Figure~\ref{fig:app-extraction}): a content classification stage (GPT-4.1) first filters the 1,043 sections, finding that approximately 60\% contain extractable fiqh rulings while the remaining 40\% are meta-commentary, philosophical discussion, or cross-references. Three models (GPT-4.1 in structured output mode, GPT-4o, and GPT-5.1) then extract in parallel, producing typed records with per-madhab rulings, legal classifications, structural components (pillars, conditions, obligations), and evidence chains. Ensemble voting via fuzzy-matching alignment resolves disagreements by majority agreement. A separate reasoning model (o1) performs adversarial verification  against the source text, checking madhab attribution, ruling content, numerical counts, and uniqueness claims: 85\% of extractions pass directly, 12\% enter iterative refinement via o3, and 3\% are rejected outright. The corpus maintains near-uniform madhab representation (approximately 25\% per school).

From this corpus, we generate 4,000 questions across five types (Section~\ref{sec:fiqh}; Figure~\ref{fig:app-qgen}). Distractor selection varies by type: for ruling identification, distractors are rulings from other schools or plausible variations; for bias detection, all options are correct (one per school); for multi-hop reasoning, distractors are partial-reasoning answers. Answer option order is randomised across all question types.

\paragraph{Quality Verification.} Each question undergoes automatic verification against five criteria: source traceability, madhab accuracy, linguistic correctness, distractor validity, and coverage balance. Additionally, 213 stratified question samples with their answers were manually reviewed by a professor in comparative jurisprudence at Al-Azhar University. The reviewer approved 207 of 213 questions (97.2\%), with three requiring minor revisions and three rejected for factual errors (two madhab misattributions, one insufficient source grounding). While the use of a single reviewer limits generalisability, the high approval rate across all five question types provides reasonable confidence in question quality.

Figure~\ref{fig:app-pipelines} presents the two core pipeline architectures for the Fiqh track: (a)~the multi-agent extraction pipeline that converts al-Jaziri's source text into a structured corpus, and (b)~the question generation pipeline that produces the 4,000 benchmark questions from that corpus.

\section{Fiqh Question Design and Bias Methodology}
\label{sec:fiqh}

The 3,200 knowledge questions and 800 bias detection questions span eight fiqh categories proportional to the source corpus: prayer (34\%), family law (20\%), ethics (16\%), \textit{tahara} (12\%), \textit{hajj} (6\%), criminal law (5\%), fasting (5\%), and zakat (3\%). Five question types test distinct facets of Islamic legal reasoning:

\paragraph{Madhab Ruling Identification (820 questions).} Given a fiqh topic and a specified madhab, select the correct ruling from four options. Distractors are populated with rulings from other schools or plausible variations. Example: \textit{``What is the Shafi'i position on congregational prayer for the five obligatory prayers?''} Correct: Communal obligation (\textit{fard kifaya}). Distractors: Maliki (recommended), Hanbali (individual obligation).

\paragraph{Multi-Hop Reasoning (810 questions).} Questions requiring synthesis of multiple corpus facts to reach the correct answer. For example, determining whether an act invalidates prayer may require knowing both the ruling's classification and its structural role (pillar vs.\ condition).

\paragraph{Madhab Bias Detection (800 questions).} All four answer options are valid, each representing the correct ruling from a different madhab. The question avoids any school specification, asking simply ``What is the ruling on [topic]?'' A model's selection reveals which school it defaults to when unconstrained.

\paragraph{Component Parsing (790 questions).} Questions asking models to classify sub-actions as pillars (\textit{arkan}), conditions of validity (\textit{shurut sihha}), conditions of obligation (\textit{shurut wujub}), or recommended acts (\textit{sunan}).

\paragraph{Comparative Fiqh (780 questions).} Questions requiring identification of points of agreement or divergence across schools.  Representative examples for all question types appear in Appendix~\ref{app:examples}

\subsection{Madhab Bias Detection Methodology}

The bias detection task exploits the inherent pluralism of Sunni jurisprudence. Because all four schools are equally valid within the tradition, no answer is ``wrong'' in the conventional sense. When a model selects an option, it implicitly endorses that school's position. By aggregating responses across 800 questions, we measure whether models exhibit systematic preferences. A model with no madhab bias would show approximately 25\% selection rate for each school; significant deviation, measured via chi-squared test against the uniform distribution, indicates implicit preference. Answer order is randomised across questions to prevent positional bias.

This extends bias measurement from the inter-cultural axis studied by \citet{naous2024beer} (Western vs.\ Islamic norms) to the intra-tradition axis (preferences among equally legitimate Sunni schools). The distinction matters practically: a user following the Maliki school who receives Hanbali-biased responses may adopt rulings their own tradition does not endorse.

\section{Ranking LLMs according to IslamicMMLU}
\label{sec:results}

\subsection{Evaluation Protocol}

We evaluate 26 models across four tiers: frontier (10 models including Gemini 3, GPT-5 series, Claude 4.5), mid-tier (8 models including GPT-4o, Llama 4, DeepSeek v3), Arabic-specialised (5 models: Fanar-Sadiq, Fanar, Fanar-C-2-27B, Jais-2-70B, ALLaM-7B), and baselines (3 models including GPT-4 and GPT-3.5-turbo); full model specifications appear in Appendix~\ref{app:models}. All models receive identical Arabic prompts (Appendix~\ref{app:prompts}) with zero-shot instructions, temperature set to 0 for deterministic output, and single-letter response format (A, B, C, or D). Invalid responses are marked incorrect.

The overall score is the equally weighted average of a model's accuracy across the three tracks. Equal weighting reflects the design principle that each track represents a distinct Islamic knowledge domain of equal scholarly importance, regardless of question count. We note that alternative weighting schemes (e.g., proportional to question count) would produce slightly different rankings; per-track scores are reported for researchers preferring different aggregation methods. For the Fiqh track, models are evaluated on the 3,200 knowledge questions; the 800 bias detection questions are scored separately as they have no single correct answer. The random baseline for 4-choice MCQ is 25\%.

\paragraph{Statistical Methodology.} Because all models are evaluated with temperature 0 (deterministic), run-to-run variance is zero and repeated trials are unnecessary. However, accuracy on a finite test set has inherent sampling uncertainty. We compute bootstrap confidence intervals (1,000 resamples of question-level correct/incorrect vectors) for all reported accuracies. McNemar's test on question-level agreement can be applied to the question-level data for pairwise comparisons; as an example, the difference between Gemini 3 Flash (93.8\%) and Gemini 3 Pro (92.3\%) is statistically significant (McNemar's $\chi^2{=}12.4$, $p{<}0.001$), confirming that the 1.5-point gap is not attributable to sampling variation.

\subsection{Overall Results}

\begin{figure*}[t]
    \centering
    \vspace{-3.5em}

    \begin{subfigure}[b]{0.46\textwidth}
        \hspace{-1.2cm}\begin{tikzpicture}
        \begin{axis}[
            width=5.5cm,
            height=8.5cm,
            scale only axis,
            xmin=20, xmax=100,
            xlabel={\tiny\rmfamily Accuracy (\%)},
            xticklabel style={font=\tiny\rmfamily},
            xlabel style={at={(axis description cs:0.5,-0.07)}},
            ytick={0,1,...,25},
            y dir=reverse,
            yticklabels={
                Gem.\ 3 Flash (93.8),
                Gem.\ 3 Pro (92.3),
                Gem.\ 2.5 Pro (90.3),
                GPT-5 (89.9),
                GPT-5.2 (89.5),
                GPT-5.1 (88.4),
                Cl.\ Son.\ 4.5 (86.2),
                Cl.\ Opus 4.5 (86.0),
                Gem.\ 2.5 Flash (83.1),
                Cl.\ 3.7 Son.\ (82.3),
                {\textbf{Fanar-Sadiq}} (81.6),
                GPT-4o (77.3),
                GPT-4.1 (75.6),
                Grok 4.1 (72.9),
                DeepSeek v3.2 (68.5),
                Ll.\ 4 Mav.\ (66.7),
                {\textbf{Fanar}} (66.0),
                Cl.\ Haiku 4.5 (65.4),
                DeepSeek v3.1 (65.4),
                {\textbf{Jais-2-70B}} (63.4),
                Ll.\ 4 Scout (59.6),
                GPT-4 (59.6),
                {\textbf{ALLaM-7B}} (59.5),
                Cl.\ 3.5 Haiku (57.5),
                {\textbf{Fanar-C}} (54.0),
                GPT-3.5-turbo (39.8)
            },
            ytick style={draw=none},
            yticklabel style={anchor=east, font=\tiny\rmfamily},
            enlarge y limits={abs=0.4},
            axis lines*=left,
            tick align=outside,
            xmajorgrids=true,
            grid style={gray!15, line width=0.3pt},
            legend style={at={(0.02,1.0)}, anchor=north west, font=\tiny\rmfamily,
                legend columns=1, draw=gray!30, fill=white, fill opacity=0.95,
                row sep=0pt, column sep=2pt},
        ]
        \draw[gray!35, line width=0.6pt] (axis cs:89.06,0) -- (axis cs:99.25,0);
        \draw[gray!35, line width=0.6pt] (axis cs:87.92,1) -- (axis cs:97.32,1);
        \draw[gray!35, line width=0.6pt] (axis cs:85.53,2) -- (axis cs:95.83,2);
        \draw[gray!35, line width=0.6pt] (axis cs:83.52,3) -- (axis cs:98.01,3);
        \draw[gray!35, line width=0.6pt] (axis cs:85.91,4) -- (axis cs:94.78,4);
        \draw[gray!35, line width=0.6pt] (axis cs:81.51,5) -- (axis cs:95.28,5);
        \draw[gray!35, line width=0.6pt] (axis cs:84.91,6) -- (axis cs:88.18,6);
        \draw[gray!35, line width=0.6pt] (axis cs:72.70,7) -- (axis cs:94.59,7);
        \draw[gray!35, line width=0.6pt] (axis cs:82.36,8) -- (axis cs:83.82,8);
        \draw[gray!35, line width=0.6pt] (axis cs:81.02,9) -- (axis cs:84.87,9);
        \draw[gray!35, line width=0.6pt] (axis cs:72.88,10) -- (axis cs:94.44,10);
        \draw[gray!35, line width=0.6pt] (axis cs:71.34,11) -- (axis cs:81.97,11);
        \draw[gray!35, line width=0.6pt] (axis cs:67.96,12) -- (axis cs:81.32,12);
        \draw[gray!35, line width=0.6pt] (axis cs:65.62,13) -- (axis cs:79.09,13);
        \draw[gray!35, line width=0.6pt] (axis cs:60.06,14) -- (axis cs:74.34,14);
        \draw[gray!35, line width=0.6pt] (axis cs:56.28,15) -- (axis cs:73.52,15);
        \draw[gray!35, line width=0.6pt] (axis cs:56.93,16) -- (axis cs:74.47,16);
        \draw[gray!35, line width=0.6pt] (axis cs:53.75,17) -- (axis cs:73.07,17);
        \draw[gray!35, line width=0.6pt] (axis cs:55.34,18) -- (axis cs:71.02,18);
        \draw[gray!35, line width=0.6pt] (axis cs:48.09,19) -- (axis cs:77.84,19);
        \draw[gray!35, line width=0.6pt] (axis cs:43.27,20) -- (axis cs:70.57,20);
        \draw[gray!35, line width=0.6pt] (axis cs:42.87,21) -- (axis cs:68.37,21);
        \draw[gray!35, line width=0.6pt] (axis cs:43.22,22) -- (axis cs:70.89,22);
        \draw[gray!35, line width=0.6pt] (axis cs:47.34,23) -- (axis cs:62.76,23);
        \draw[gray!35, line width=0.6pt] (axis cs:41.85,24) -- (axis cs:60.72,24);
        \draw[gray!35, line width=0.6pt] (axis cs:32.44,25) -- (axis cs:45.16,25);
        \addplot[only marks, mark=*, mark size=2pt, trackquran] coordinates {
            (99.25,0) (97.32,1) (95.83,2) (98.01,3) (94.78,4) (95.28,5)
            (88.18,6) (94.59,7) (82.36,8) (81.02,9) (94.44,10)
            (71.34,11) (67.96,12) (79.09,13) (60.06,14) (56.28,15)
            (56.93,16) (53.75,17) (55.34,18) (48.09,19)
            (43.27,20) (42.87,21) (43.22,22) (47.34,23) (41.85,24) (32.44,25)
        };
        \addplot[only marks, mark=square*, mark size=1.5pt, trackhadith] coordinates {
            (93.00,0) (91.72,1) (89.67,2) (88.22,3) (87.77,4) (88.32,5)
            (85.37,6) (90.80,7) (83.82,8) (84.87,9) (72.88,10)
            (81.97,11) (81.32,12) (65.62,13) (71.14,14) (73.52,15)
            (66.51,16) (73.07,17) (71.02,18) (77.84,19)
            (64.87,20) (68.37,21) (70.89,22) (62.76,23) (60.72,24) (41.64,25)
        };
        \addplot[only marks, mark=triangle*, mark size=2pt, trackfiqh] coordinates {
            (89.06,0) (87.92,1) (85.53,2) (83.52,3) (85.91,4) (81.51,5)
            (84.91,6) (72.70,7) (83.14,8) (81.13,9) (77.36,10)
            (78.49,11) (77.48,12) (73.96,13) (74.34,14) (70.31,15)
            (74.47,16) (69.43,17) (69.81,18) (64.40,19)
            (70.57,20) (67.42,21) (64.53,22) (62.39,23) (59.53,24) (45.16,25)
        };
        \draw[dashed, gray!50, line width=0.8pt] (axis cs:25,{-0.8}) -- (axis cs:25,26.2);
        \node[font=\tiny, gray!50, anchor=south, rotate=90] at (axis cs:25,23) {25\% random};
        \draw[gray!25, line width=0.3pt, dashed] (axis cs:20,9.5) -- (axis cs:100,9.5);
        \legend{Quran, Hadith, Fiqh}
        \end{axis}
        \end{tikzpicture}
        \caption{Per-track accuracy for all 26 models sorted by overall score. Arabic-specific models in \textbf{bold}.}
        \label{fig:overall-results}
    \end{subfigure}%
    \hspace{0.04\textwidth}%
    \begin{subfigure}[b]{0.46\textwidth}
        \centering
        \begin{tikzpicture}[
          mlbl/.style={font=\tiny, anchor=east, text=black!90},
        ]
          \def\xs{0.13}
          \def\ys{0.35}
          \def\dotrad{1.3pt}

          \newcommand{\biasrow}[6]{%
            \pgfmathsetmacro{\yy}{#1*\ys}%
            \pgfmathsetmacro{\xmin}{min(min(#3,#4),min(#5,#6))}%
            \pgfmathsetmacro{\xmax}{max(max(#3,#4),max(#5,#6))}%
            \draw[line width=0.3pt, black!25] (\xmin*\xs, \yy) -- (\xmax*\xs, \yy);%
            \fill[madhabHanafi] (#3*\xs, \yy) circle (\dotrad);%
            \fill[madhabMaliki] (#4*\xs, \yy) circle (\dotrad);%
            \fill[madhabShafii] (#5*\xs, \yy) circle (\dotrad);%
            \fill[madhabHanbali] (#6*\xs, \yy) circle (\dotrad);%
            \node[mlbl] at (-13*\xs-0.08, \yy) {#2};%
          }

          \fill[tierBaseline!6] (-13*\xs-2.1, -0.15) rectangle (24*\xs+0.1, 3.5*\ys);
          \fill[tierArabic!6] (-13*\xs-2.1, 3.5*\ys) rectangle (24*\xs+0.1, 7.5*\ys);
          \fill[tierMid!6] (-13*\xs-2.1, 7.5*\ys) rectangle (24*\xs+0.1, 15.5*\ys);
          \fill[tierFrontier!6] (-13*\xs-2.1, 15.5*\ys) rectangle (24*\xs+0.1, 25*\ys+0.15);

          \draw[dashed, line width=0.5pt, black!50] (0, -0.2) -- (0, 25*\ys+0.15);
          \node[font=\tiny, black!60, above] at (0, 25*\ys+0.15) {0};

          \foreach \d in {-10,-5,5,10,15,20} {
            \draw[line width=0.2pt, black!12] (\d*\xs, -0.2) -- (\d*\xs, 25*\ys+0.15);
          }

          \foreach \d in {-10,-5,0,5,10,15,20} {
            \node[font=\tiny, black!70, below] at (\d*\xs, -0.3) {\d};
          }
          \node[font=\tiny, black!80] at (2*\xs, -0.8) {deviation from 25\% ($\Delta$\%)};

          \biasrow{0}{GPT-3.5-turbo}{10.63}{-0.48}{-2.01}{-8.14}
          \biasrow{1}{Fanar-C}{22.40}{-6.50}{-9.39}{-6.50}
          \biasrow{2}{Cl.\ 3.5 Haiku}{7.95}{7.57}{-2.78}{-12.74}
          \biasrow{3}{GPT-4}{8.33}{2.97}{-5.08}{-6.23}
          \biasrow{4}{Fanar}{7.18}{2.59}{-4.31}{-5.46}
          \biasrow{5}{ALLaM-7B}{5.27}{-3.16}{-3.16}{1.05}
          \biasrow{6}{Jais-2-70B}{8.10}{-8.62}{0.78}{-0.26}
          \biasrow{7}{Fanar-Sadiq}{7.57}{-0.10}{-3.16}{-4.31}
          \biasrow{8}{Cl.\ Haiku 4.5}{5.27}{0.67}{1.05}{-6.99}
          \biasrow{9}{DeepSeek v3.1}{11.78}{4.50}{-5.08}{-11.21}
          \biasrow{10}{Ll.\ 4 Mav.}{0.67}{2.97}{3.35}{-6.99}
          \biasrow{11}{Ll.\ 4 Scout}{-2.01}{5.27}{0.29}{-3.54}
          \biasrow{12}{Grok 4.1}{6.03}{-1.25}{-0.48}{-4.31}
          \biasrow{13}{DeepSeek v3.2}{7.18}{0.67}{-0.10}{-7.76}
          \biasrow{14}{GPT-4.1}{2.59}{-2.78}{-0.86}{1.05}
          \biasrow{15}{GPT-4o}{9.10}{-5.46}{-5.46}{1.82}
          \biasrow{16}{Cl.\ Opus 4.5}{17.15}{-5.08}{-3.93}{-8.14}
          \biasrow{17}{Cl.\ 3.7 Son.}{-1.25}{1.44}{5.27}{-5.46}
          \biasrow{18}{GPT-5.1}{-10.44}{-3.54}{3.35}{10.63}
          \biasrow{19}{Gem.\ 2.5 Flash}{-0.86}{-3.54}{0.67}{3.74}
          \biasrow{20}{GPT-5}{-3.54}{-2.78}{3.74}{2.59}
          \biasrow{21}{Cl.\ Son.\ 4.5}{1.05}{4.50}{-2.78}{-2.78}
          \biasrow{22}{Gem.\ 2.5 Pro}{-0.48}{-0.10}{2.20}{-1.63}
          \biasrow{23}{GPT-5.2}{-6.99}{-1.63}{5.65}{2.97}
          \biasrow{24}{Gem.\ 3 Pro}{4.50}{-3.93}{2.97}{-3.54}
          \biasrow{25}{Gem.\ 3 Flash}{-0.48}{1.82}{-2.01}{0.67}

          \node[font=\tiny, black!55, rotate=90, anchor=south]
            at (24*\xs+0.3, 1.5*\ys) {baseline};
          \node[font=\tiny, black!55, rotate=90, anchor=south]
            at (24*\xs+0.3, 5.5*\ys) {arabic};
          \node[font=\tiny, black!55, rotate=90, anchor=south]
            at (24*\xs+0.3, 11.5*\ys) {mid-tier};
          \node[font=\tiny, black!55, rotate=90, anchor=south]
            at (24*\xs+0.3, 20.5*\ys) {frontier};

          \node[draw=black!55, fill=white, rounded corners=1pt,
            font=\tiny, inner sep=3pt, anchor=north east]
            at (24*\xs+0.4, 25*\ys+0.55) {%
            \tikz[baseline=-0.5ex]{\fill[madhabHanafi] (0,0) circle (1.2pt);} Han.\hspace{3pt}%
            \tikz[baseline=-0.5ex]{\fill[madhabMaliki] (0,0) circle (1.2pt);} Mal.\hspace{3pt}%
            \tikz[baseline=-0.5ex]{\fill[madhabShafii] (0,0) circle (1.2pt);} Sha.\hspace{3pt}%
            \tikz[baseline=-0.5ex]{\fill[madhabHanbali] (0,0) circle (1.2pt);} Hba.%
          };

        \end{tikzpicture}
        \caption{Madhab selection deviation from uniform 25\% on 800 bias detection questions.}
        \label{fig:madhab-bias}
    \end{subfigure}
    \caption{Model evaluation results. (a) Per-track accuracy with cross-track spread. (b) Madhab bias: dots near the centre line indicate balanced school selection.}
    \label{fig:main-figure}

    \vspace*{-1.5em}

\end{figure*}
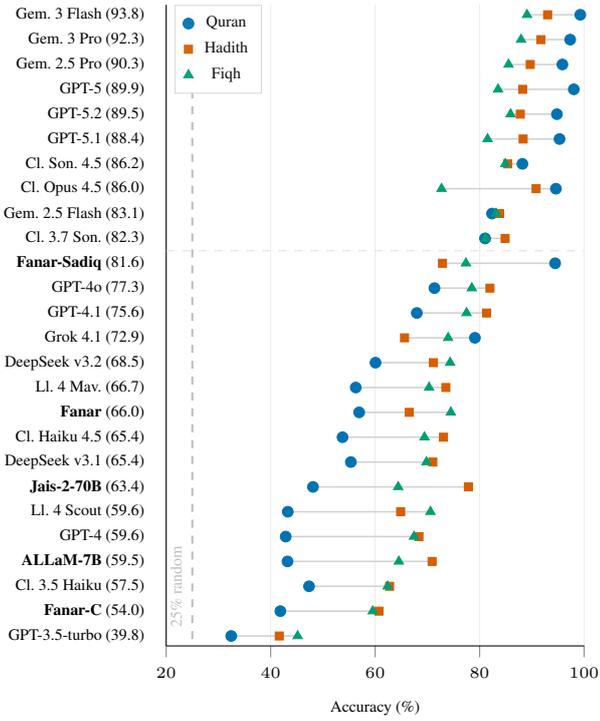
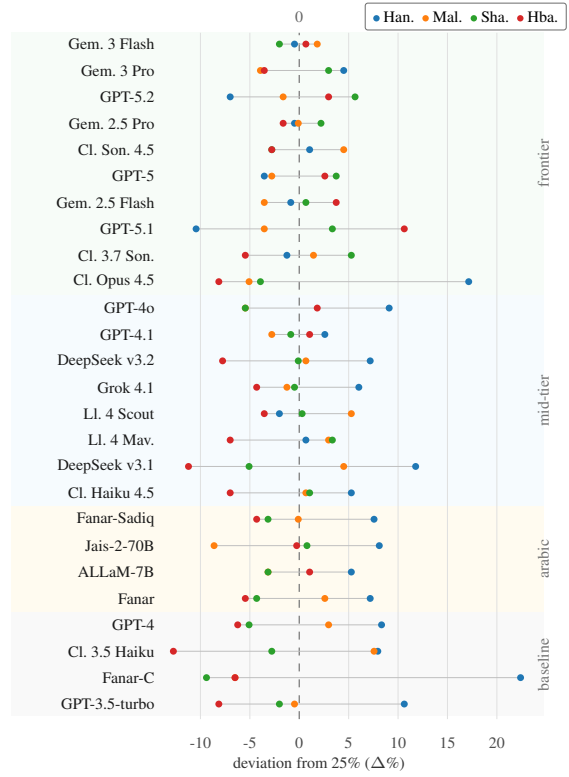

Table~\ref{tab:main-results} presents accuracy for all 26 evaluated models across all three tracks. Gemini 3 Flash achieves the highest overall accuracy (93.77\%), followed by Gemini 3 Pro (92.32\%) and Gemini 2.5 Pro (90.34\%). Google models dominate the top three positions. GPT-5 (89.92\%) is the top non-Google model; Claude Sonnet 4.5 (86.15\%) is the top Anthropic model. The 54-point gap between the top model and the GPT-3.5-turbo baseline (39.75\%) confirms that the benchmark distinguishes  model capabilities effectively above the 25\% random baseline floor.

\begin{table}[t]
    \centering
    \footnotesize
    \setlength{\tabcolsep}{3pt}
    \begin{tabular}{lcccc}
        \toprule
        \textbf{Model} & \textbf{Overall} & \textbf{Quran} & \textbf{Hadith} & \textbf{Fiqh} \\
        \midrule
        Gemini 3 Flash    & \textbf{93.8} & \textbf{99.3} & 93.0 & \textbf{89.1} \\
        Gemini 3 Pro      & 92.3 & 97.3 & 91.7 & 87.9 \\
        Gemini 2.5 Pro    & 90.3 & 95.8 & 89.7 & 85.5 \\
        GPT-5             & 89.9 & 98.0 & 88.2 & 83.5 \\
        GPT-5.2           & 89.5 & 94.8 & 87.8 & 85.9 \\
        GPT-5.1           & 88.4 & 95.3 & 88.3 & 81.5 \\
        Cl.\ Sonnet 4.5   & 86.2 & 88.2 & 85.4 & 84.9 \\
        Cl.\ Opus 4.5     & 86.0 & 94.6 & \textbf{90.8} & 72.7 \\
        Gem.\ 2.5 Flash   & 83.1 & 82.4 & 83.8 & 83.1 \\
        Cl.\ 3.7 Sonnet   & 82.3 & 81.0 & 84.9 & 81.1 \\
        \midrule
        \textbf{Fanar-Sadiq}       & 81.6 & 94.4 & 72.9 & 77.4 \\
        GPT-4o            & 77.3 & 71.3 & 82.0 & 78.5 \\
        GPT-4.1           & 75.6 & 68.0 & 81.3 & 77.5 \\
        Grok 4.1 Fast     & 72.9 & 79.1 & 65.6 & 74.0 \\
        DeepSeek v3.2     & 68.5 & 60.1 & 71.1 & 74.3 \\
        Ll.\ 4 Maverick   & 66.7 & 56.3 & 73.5 & 70.3 \\
        \textbf{Fanar}             & 66.0 & 56.9 & 66.5 & 74.5 \\
        Cl.\ Haiku 4.5    & 65.4 & 53.8 & 73.1 & 69.4 \\
        DeepSeek v3.1     & 65.4 & 55.3 & 71.0 & 69.8 \\
        \textbf{Jais-2-70B}       & 63.4 & 48.1 & 77.8 & 64.4 \\
        Ll.\ 4 Scout      & 59.6 & 43.3 & 64.9 & 70.6 \\
        GPT-4             & 59.6 & 42.9 & 68.4 & 67.4 \\
        \textbf{ALLaM-7B}         & 59.5 & 43.2 & 70.9 & 64.5 \\
        Cl.\ 3.5 Haiku    & 57.5 & 47.3 & 62.8 & 62.4 \\
        Fanar-C-2-27B     & 54.0 & 41.9 & 60.7 & 59.5 \\
        \midrule
        GPT-3.5-turbo     & 39.8 & 32.4 & 41.6 & 45.2 \\
        \bottomrule
    \end{tabular}
    \caption{IslamicMMLU accuracy (\%) for all 26 models. Arabic-specific models in \textbf{bold}. Fiqh accuracy is on the 3,200 knowledge questions (800 bias detection questions scored separately). Per-question-type breakdowns are in Table~\ref{tab:expanded}.}
    \label{tab:main-results}
\end{table}

\subsection{Per-Track Analysis}

Performance varies substantially across tracks (per-task breakdowns in Table~\ref{tab:expanded}). The Quran track has the widest spread: 99.25\% (Gemini 3 Flash) to 32.44\% (GPT-3.5-turbo), a 66.8-point gap. Four models exceed 95\%, yet weaker models score well below 50\%, making it the most discriminative domain. The Hadith track spans 93.00\% to 41.64\% (51.4 points), and the Fiqh track spans 89.06\% to 45.16\% (43.9 points).

Cross-track discrepancies reveal model-specific strengths. Claude Opus 4.5 achieves strong Quran (94.6\%) and Hadith (90.8\%) scores but notably weaker Fiqh (72.7\%), suggesting difficulty with comparative reasoning. Fanar-Sadiq scores 94.4\% on Quran (competitive with frontier models) but only 72.9\% on Hadith, indicating uneven domain coverage. These patterns demonstrate the value of multi-track evaluation: a single aggregate score obscures important domain-specific capabilities and weaknesses.

\subsection{Arabic-Specific Models}

Arabic-specialised models show mixed results. Fanar-Sadiq~\citep{fanar2025} (81.56\%) performs competitively, likely benefiting from its specialised Islamic retrieval-augmented generation system. Jais-2-70B (63.4\%) and ALLaM-7B (59.5\%) rank in the lower-mid tier despite their large-scale Arabic pretraining, with both showing notably strong Hadith performance (77.8\% and 70.9\%) but weak Quran scores (48.1\% and 43.2\%). Fanar (66.0\%) and Fanar-C-2-27B (54.0\%) score lower still, clustering with or below general-purpose mid-tier models despite their Arabic-specific training. Arabic pretraining alone appears insufficient for Islamic knowledge; domain-specific training data and retrieval augmentation matter more than language coverage.

\subsection{Madhab Bias Analysis}

For the 800 Fiqh bias detection questions, we measure each model's selection distribution across the four schools. Figure~\ref{fig:madhab-bias} visualises the deviation from uniform 25\% selection for all 26 models.

\paragraph{Variable Bias.} Unlike cultural bias studies showing consistent Western skew~\citep{naous2024beer}, madhab preferences are model-specific. Gemini 3 Flash shows near-uniform distribution ($\chi^2{=}0.32$, $p{>}0.9$), while GPT-5.1 significantly favours Hanbali (35.6\%). Arabic-specific models show moderate Hanafi preference (32--33\%), possibly reflecting Gulf training data where Hanafi jurisprudence predominates. With 26 independent tests, we apply Bonferroni correction ($\alpha_{\text{adj}} = 0.05/26 = 0.0019$); only Claude Opus 4.5 ($\chi^2{=}16.06$) retains significance after correction.

\paragraph{Bias--Accuracy Relationship.} A moderate negative correlation exists between accuracy and bias magnitude (Pearson $r{=}{-}0.38$, $p{=}0.049$; Spearman $\rho{=}{-}0.53$, $p{=}0.005$), suggesting higher-performing models tend toward more balanced school selection. However, notable exceptions exist (some high-accuracy models show elevated bias), indicating that capability improvements alone do not guarantee fairness within traditions.

\subsection{Error Analysis}

\paragraph{Fiqh.} Qualitative examination of Fiqh errors reveals three recurring patterns. \textit{Madhab confusion}: models return rulings from an incorrect school when a specific madhab is specified, indicating difficulty with school-specific retrieval. \textit{Multi-hop reasoning failure}: questions requiring synthesis of multiple facts (e.g., determining consequences of omitting a pillar vs.\ violating a condition) show the largest frontier-to-baseline gap (87.2\% vs.\ 41.1\%). \textit{Comparative difficulty}: Comparative Fiqh questions achieve the lowest top-model accuracy (76.7\%), as cross-school synthesis demands tracking divergent positions. We note that per-type accuracy differences should be interpreted with caution for Comparative Fiqh ($n{=}780$) and Component Parsing ($n{=}790$), where smaller question pools yield wider confidence intervals than the larger Ruling Identification ($n{=}820$) and Multi-Hop ($n{=}810$) sets.

\paragraph{Quran.} The wide performance spread on this track suggests Quranic textual knowledge is highly discriminative: models either have extensive exposure to Quranic content or not  (see Appendix~\ref{app:contamination} for contamination considerations). Ayah identification is the easiest subtype for top models, while surah identification based on thematic attributes proves harder, requiring reasoning beyond surface-level text matching. Weaker models frequently confuse surahs of similar length or revelation period.

\paragraph{Hadith.} Chapter classification is the hardest task, where mapping narrative content to taxonomic chapter labels requires semantic understanding beyond keyword matching.

\section{Public Leaderboard}
\label{sec:leaderboard}

To support future evaluation as new models are released, we deploy IslamicMMLU as an interactive leaderboard on HuggingFace.\footnote{\url{https://huggingface.co/spaces/islamicmmlu/leaderboard}} The platform enables evaluation of models by providing an API key and model identifier; the platform runs the full pipeline automatically and publishes results to a public dataset.

All 10,013 questions follow a standardised four-option MCQ format with consistent metadata (track, category, question type, difficulty level). This standardised structure is designed for extensibility: additional Islamic knowledge domains -- such as \textit{aqidah} (creed), \textit{sirah} (prophetic biography), or \textit{tafsir} (Quranic exegesis) -- can be added as new tracks without modifying the evaluation infrastructure. The leaderboard provides rankings, analytics including madhab bias visualisations, and side-by-side model comparison across all tracks and question types.

\section{Conclusion and Future Work}
\label{sec:conclusion}

We introduced IslamicMMLU, a comprehensive benchmark of 10,013 questions across 12 tasks covering the Quran, Hadith, and Fiqh. Our evaluation of 26 LLMs demonstrates the benchmark’s strong discriminative power, revealing a 54-point performance gap between frontier and legacy models. Notably, we find a significant correlation between model capability and reduced madhab bias, suggesting that advanced reasoning may naturally mitigate intra-tradition skew.

Future work will expand this framework by incorporating Shia jurisprudence to ensure broader representation. Furthermore, we aim to extend the MMLU paradigm to additional domains. By releasing our evaluation suite and public leaderboard, we provide a standardized foundation for the culturally aware and technically rigorous evaluation of LLMs within Islamic scholarship.

\section{Limitations}
\label{sec:limitations}

\paragraph{Sunni Scope.} All three tracks focus on Sunni Islam. The Quran and Hadith tracks use Sunni canonical sources; the Fiqh track covers only the four Sunni madhahib. Shia perspectives (Ja'fari, Zaydi, Ismaili) are excluded, limiting applicability to approximately 15\% of the global Muslim population. Extending the benchmark to cover Shia schools is planned for future work.

\paragraph{Single Source Dependency (Fiqh).} The Fiqh corpus derives entirely from al-Jaziri's encyclopedia. While authoritative, this limits coverage of minority opinions within schools, regional variations, and contemporary \textit{ijtihad} (independent legal reasoning).

\paragraph{Partial Human Validation.} The Fiqh track has documented external expert validation on a stratified sample (207/213 approved, 97.2\%). We acknowledge that inter-annotator agreement from multiple experts would strengthen validation; the current 97.2\% approval rate from a single domain expert provides initial quality evidence. Validation processes for the Quran and Hadith tracks are documented in their respective source works but not independently re-validated for IslamicMMLU.

\paragraph{Arabic-Only.} All questions are in Modern Standard Arabic. This excludes cross-lingual evaluation relevant for diaspora communities and multilingual Islamic scholarship.

\paragraph{MCQ Format.} Multiple-choice format measures recognition rather than generation. A model selecting the correct ruling from four options may not produce that ruling unprompted. The four-option format also permits a 25\% baseline through random guessing.

\paragraph{Temporal Validity.} Results represent model capabilities at evaluation time (January--February 2026). Rankings and bias patterns may shift with model updates.

\section{Ethics Statement}
\label{sec:ethics}

\paragraph{Religious Content Sensitivity.} We handle Islamic content with scholarly respect, drawing exclusively from established academic sources. Our framing maintains neutrality across madhahib, and we do not position any school as preferred. The benchmark evaluates model knowledge, not religious truth claims. Results from the Hadith authenticity grading task should not be treated as authoritative religious judgements.

\paragraph{Privacy.} The dataset contains no personally identifiable information. All content derives from published scholarly works.

\paragraph{Potential for Misuse.} Bias detection results could be misrepresented to claim models are ``anti-Islamic'' or favour particular sects. We emphasise: variable bias likely reflects training data composition, not intentional design; bias measurements are probabilistic tendencies, not deterministic behaviours; and findings should not be used to make inflammatory claims about AI companies or religious communities.

\paragraph{Deployment.} We discourage using IslamicMMLU results to market AI systems as authoritative Islamic advisors. Religious guidance should involve qualified human scholars.

\paragraph{Positionality.} This benchmark was developed within an Islamic studies research group at a UK university. The primary author's background in Islamic knowledge informed the Fiqh track design. The benchmark focuses on Sunni tradition, and we acknowledge this scope limitation.

\bibliography{references}

\appendix

\section{Model Specifications}
\label{app:models}

All models were accessed via API between December 2025 and March 2026. Frontier models: Gemini 3 Flash/Pro, Gemini 2.5 Flash/Pro (Google); GPT-5, GPT-5.1, GPT-5.2 (OpenAI); Claude Sonnet 4.5, Claude 3.7 Sonnet, Claude Opus 4.5 (Anthropic). Mid-tier: GPT-4o, GPT-4.1 (OpenAI); Claude Haiku 4.5 (Anthropic); DeepSeek v3.1, v3.2; Llama 4 Scout, Maverick (Meta); Grok 4.1 Fast (xAI). Arabic-specific: Fanar-Sadiq, Fanar, Fanar-C-2-27B (QCRI); Jais-2-70B (Inception/MBZUAI); ALLaM-7B (SDAIA). Baselines: GPT-4, Claude 3.5 Haiku, GPT-3.5-turbo.

\section{Evaluation Prompts}
\label{app:prompts}

All models received Arabic prompts:

\begin{quote}
    \footnotesize
    \textit{``Ajib `an al-su'al al-tali bi-ikhtiyar al-ijaba al-sahiha:''} (Answer the following question by selecting the correct answer:)

    \textit{[Question text in Arabic]}

    \textit{A)} [Option A] \quad \textit{B)} [Option B]\\
    \textit{C)} [Option C] \quad \textit{D)} [Option D]

    \textit{``Ajib bi-l-harf faqat.''}\\(Answer with the letter only.)
\end{quote}

Parameters: temperature 0, max tokens 10, no system prompt.

\onecolumn
\section{Expanded Results by Question Type}
\label{app:expanded}

\begin{table}[h]
    \centering
    \tiny
    \setlength{\tabcolsep}{2pt}
    \begin{tabular}{l|ccc|cccc|ccccc}
        \toprule
        & \multicolumn{3}{c|}{\textbf{Quran}} & \multicolumn{4}{c|}{\textbf{Hadith}} & \multicolumn{5}{c}{\textbf{Fiqh (by madhab)}} \\
        \textbf{Model} & \textbf{Ayah} & \textbf{Surah} & \textbf{Count} & \textbf{Source} & \textbf{Cloze} & \textbf{Chap.} & \textbf{Auth.} & \textbf{Han.} & \textbf{Mal.} & \textbf{Sha.} & \textbf{Hba.} & \textbf{Comp.} \\
        \midrule
        Gem.\ 3 Flash    & 99.1 & 99.4 & 100  & 96.4 & 95.8 & 96.9 & 82.9 & 89.7 & 88.9 & 92.2 & 87.1 & 80.0 \\
        Gem.\ 3 Pro      & 96.9 & 97.5 & 100  & 93.1 & 95.7 & 95.7 & 82.4 & 92.6 & 84.9 & 90.0 & 84.8 & 82.9 \\
        Gem.\ 2.5 Pro    & 96.6 & 94.2 & 100  & 91.7 & 95.0 & 94.1 & 77.9 & 88.2 & 81.9 & 85.0 & 87.6 & 82.9 \\
        GPT-5             & 98.9 & 96.6 & 100  & 87.6 & 95.2 & 94.0 & 76.1 & 84.2 & 84.4 & 82.2 & 83.7 & 80.0 \\
        GPT-5.2           & 98.4 & 92.0 & 81.6 & 89.6 & 92.1 & 93.6 & 75.8 & 88.2 & 85.9 & 82.8 & 87.6 & 80.0 \\
        GPT-5.1           & 97.5 & 91.8 & 100  & 86.0 & 94.0 & 94.2 & 79.1 & 77.3 & 82.9 & 82.8 & 84.3 & 77.1 \\
        Cl.\ Son.\ 4.5   & 85.4 & 90.4 & 98.2 & 89.8 & 91.0 & 91.8 & 68.9 & 84.7 & 82.4 & 85.6 & 87.6 & 82.9 \\
        Cl.\ Opus 4.5    & 93.0 & 96.0 & 99.1 & 94.7 & 95.6 & 96.6 & 76.3 & 70.9 & 70.4 & 72.2 & 79.2 & 65.7 \\
        Gem.\ 2.5 Flash  & 78.1 & 86.2 & 93.9 & 89.4 & 85.3 & 88.4 & 72.2 & 84.7 & 79.9 & 87.2 & 83.1 & 71.4 \\
        Cl.\ 3.7 Son.    & 77.3 & 83.9 & 94.7 & 90.9 & 90.2 & 92.9 & 65.5 & 83.3 & 79.4 & 81.7 & 80.9 & 77.1 \\
        \midrule
        Fanar-Sadiq       & 93.2 & 95.7 & 97.4 & 62.8 & 97.6 & 83.7 & 47.4 & 71.9 & 79.4 & 79.4 & 81.5 & 65.7 \\
        GPT-4o            & 72.5 & 71.1 & 62.3 & 79.5 & 81.8 & 89.9 & 76.7 & 75.9 & 76.9 & 78.3 & 83.7 & 77.1 \\
        GPT-4.1           & 67.3 & 69.2 & 65.8 & 82.1 & 83.5 & 88.6 & 71.1 & 72.9 & 75.9 & 80.6 & 81.5 & 77.1 \\
        Grok 4.1 Fast     & 78.9 & 76.5 & 100  & 50.3 & 78.1 & 86.4 & 47.6 & 70.9 & 71.9 & 72.8 & 79.8 & 80.0 \\
        DeepSeek v3.2     & 41.5 & 78.5 & 99.1 & 72.0 & 83.3 & 88.3 & 40.9 & 69.0 & 74.4 & 75.0 & 79.2 & 77.1 \\
        Ll.\ 4 Maverick   & 46.8 & 63.1 & 94.7 & 79.9 & 75.5 & 85.1 & 53.5 & 67.5 & 71.4 & 67.2 & 74.2 & 77.1 \\
        Fanar             & 39.9 & 75.3 & 82.5 & 62.3 & 79.9 & 83.7 & 40.1 & 67.5 & 76.9 & 76.1 & 78.7 & 71.4 \\
        Cl.\ Haiku 4.5   & 44.6 & 61.3 & 84.2 & 73.9 & 73.6 & 85.4 & 59.4 & 71.4 & 66.8 & 66.7 & 71.9 & 74.3 \\
        DeepSeek v3.1     & 34.2 & 76.8 & 95.6 & 71.0 & 84.6 & 87.6 & 40.8 & 67.0 & 68.3 & 68.3 & 75.3 & 74.3 \\
        Jais-2-70B        & 41.8 & 57.3 & 39.5 & 78.8 & 78.7 & 87.0 & 66.9 & 64.0 & 60.3 & 65.6 & 69.7 & 57.1 \\
        Ll.\ 4 Scout      & 39.3 & 48.5 & 42.1 & 55.7 & 70.7 & 83.6 & 49.5 & 70.0 & 64.8 & 72.8 & 74.7 & 74.3 \\
        GPT-4             & 35.5 & 50.7 & 55.3 & 55.2 & 69.0 & 83.9 & 65.4 & 64.0 & 69.3 & 67.2 & 68.0 & 74.3 \\
        ALLaM-7B          & 31.2 & 54.9 & 70.2 & 63.1 & 77.8 & 86.3 & 56.4 & 62.6 & 62.8 & 68.3 & 63.5 & 71.4 \\
        Cl.\ 3.5 Haiku   & 40.9 & 56.7 & 39.5 & 58.7 & 71.9 & 85.6 & 34.8 & 57.6 & 59.8 & 64.4 & 66.9 & 71.4 \\
        Fanar-C-2-27B     & 34.5 & 45.8 & 79.8 & 53.8 & 70.6 & 81.1 & 36.4 & 53.0 & 63.8 & 58.2 & 64.7 & 55.6 \\
        \midrule
        GPT-3.5-turbo     & 26.6 & 40.3 & 29.8 & 37.9 & 49.8 & 63.3 & 15.5 & 43.8 & 45.7 & 44.4 & 44.9 & 54.3 \\
        \bottomrule
    \end{tabular}
    \caption{Expanded accuracy (\%) by question type per track. Quran: Ayah Identification, Surah Identification, Ayah Count. Hadith: Source Identification, Cloze Completion, Chapter Classification, Authenticity Grading. Fiqh: accuracy on questions tagged by madhab (Hanafi, Maliki, Shafi'i, Hanbali) and Comparative Fiqh cross-school questions.}
    \label{tab:expanded}
\end{table}

\section{Contamination Considerations}
\label{app:contamination}

We cannot determine whether evaluated models encountered our source texts during training. The Quran is among the most widely available Arabic texts online, which may partially explain high Quran track scores. However, several factors mitigate contamination concerns. First, our questions test understanding, not memorisation: ayah identification requires contextual attribution rather than verbatim recall. Second, performance is highly variable across models, and if questions were trivially solvable through memorisation, we would expect uniform high scores rather than the observed 66.8-point Quran spread. Third, Fiqh questions are generated from structured extractions of al-Jaziri's text and are unlikely to appear verbatim in training corpora. Fourth, Hadith isnad trimming and TF-IDF-based cloze design produce novel question formulations.

\clearpage

\section{Example Questions by Type}
\label{app:examples}
\vspace{-4pt}
\begin{table}[!ht]
\centering
\caption{Representative example for each of the 12 question types. Arabic originals with English translations; correct answers \textbf{bolded}. For bias detection, all four options are correct (one per school); school labels are hidden from the model.}
\label{tab:examples}
\includegraphics[width=\textwidth]{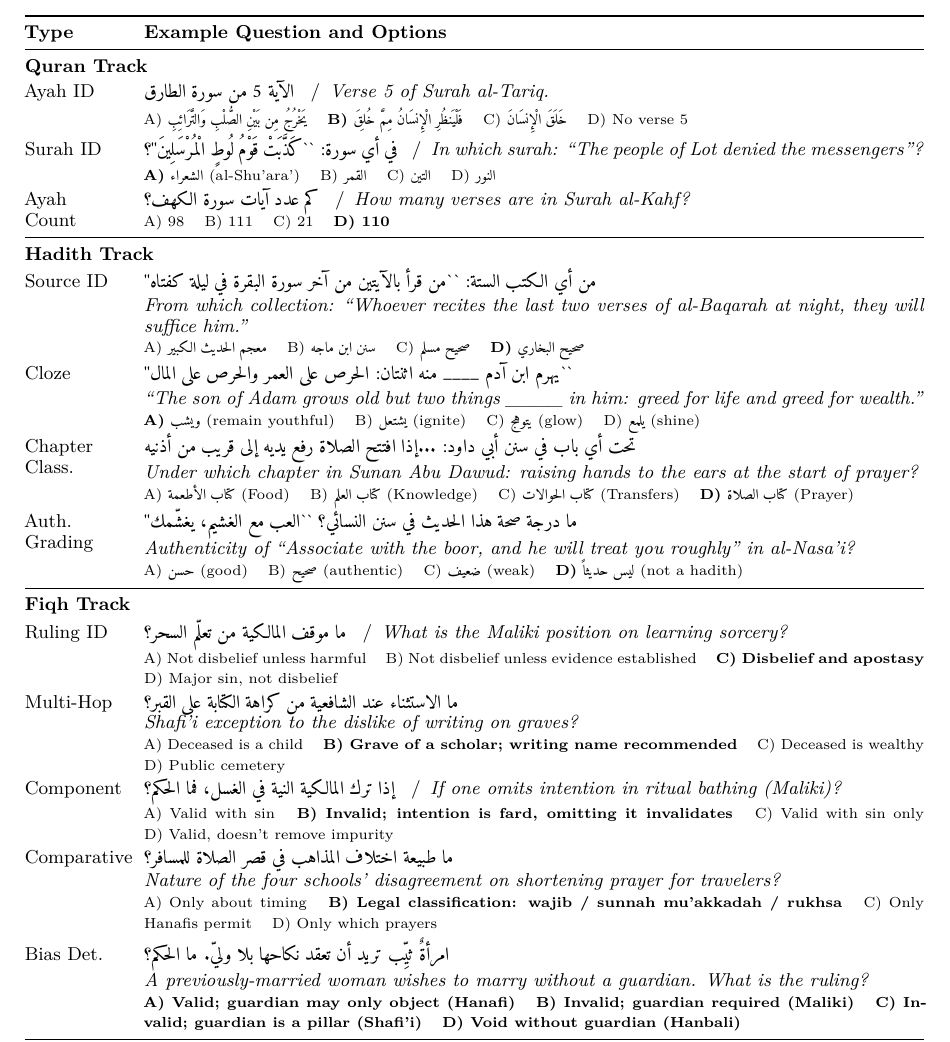}
\end{table}

\end{document}